\documentclass[runningheads]{llncs}

\usepackage[T1]{fontenc}
\usepackage{graphicx}
\usepackage{cite}
\usepackage{algorithmic}
\usepackage{textcomp}
\usepackage{array}
\usepackage{subcaption}
\usepackage{stfloats}
\usepackage{url}
\usepackage{verbatim}
\usepackage{times}
\usepackage{epsfig}
\usepackage{amsmath}
\usepackage{amssymb}
\usepackage{amsfonts}
\usepackage{booktabs}
\usepackage{fixltx2e}
\usepackage{colortbl}
\usepackage[vlined,ruled]{algorithm2e}
\usepackage{xcolor}
\usepackage{color}
\usepackage{soul}
\usepackage{multirow}
\usepackage{float}
\usepackage{balance}
\usepackage{makecell}
\usepackage{caption}
\usepackage{subcaption}
\usepackage{lipsum}
\usepackage{rotating}
\usepackage{adjustbox}
\usepackage{pdflscape}
\usepackage{siunitx}
\usepackage{lscape}
\usepackage{pifont}
\usepackage{blindtext}
\usepackage{tabu}
\usepackage[T1]{fontenc}
\usepackage[utf8]{inputenc}
\usepackage{flexisym}
\usepackage{forest}
\usepackage{longtable}
\usepackage[multiple]{footmisc}
\usepackage{ctable}
\usepackage{epstopdf}
\usepackage{fixfoot}
\usepackage{multicol}
\usepackage{tabularray}
\usepackage{tabularx}
\usepackage[most]{tcolorbox}
\usepackage{xspace}
\usepackage{academicons}
\usepackage{amsmath}
\usepackage{cuted}
\usepackage{graphicx}
\usepackage{subcaption}
\usepackage{pgfplots}
\pgfplotsset{compat=1.17}
\makeatletter
\DeclareRobustCommand\onedot{\futurelet\@let@token\@onedot}
\def\@onedot{\ifx\@let@token.\else.\null\fi\xspace}
 
\def\ie{\emph{i.e}\onedot}

\def\wrt{w.r.t\onedot} 
\def\etal{\emph{et al}\onedot}
\makeatother
\begin{document}
\title{ActNetFormer: Transformer-ResNet Hybrid Method for
Semi-Supervised Action Recognition in Videos\thanks{This research is supported by the Global Research Excellence Scholarship, Monash University, Malaysia. This research is also supported, in part, by the Global Excellence and Mobility Scholarship (GEMS), Monash University, Malaysia \& Australia.}}
\titlerunning{Transformer-ResNet Hybrid Pipeline for
Semi-Supervised Action Recognition}

%
%
\author{Sharana Dharshikgan Suresh Dass\inst{1} \and
Hrishav Bakul Barua\inst{1,2} \and
Ganesh Krishnasamy\inst{1}
Raveendran Paramesran\inst{1} \and
Raphaël C.-W. Phan\inst{1}
}
\authorrunning{Dass et al.}
%
\institute{School of Information Technology, Monash University, Malaysia \and
Robotics and Autonomous Systems Lab, TCS Research, India
\email{}\\
\url{} 
\email{\{sharana.sureshdass, hrishav.barua, ganesh.krishnasamy,\\raveendran.paramesran, raphael.phan\}@monash.edu}}
\maketitle              
\begin{abstract}
Human action or activity recognition in videos is a fundamental task in computer vision with applications in surveillance and monitoring, self-driving cars, sports analytics, human-robot interaction and many more. Traditional supervised methods require large annotated datasets for training, which are expensive and time-consuming to acquire. This work proposes a novel approach using Cross-Architecture Pseudo-Labeling with contrastive learning for semi-supervised action recognition. Our framework leverages both labeled and unlabeled data to robustly learn action representations in videos, combining pseudo-labeling with contrastive learning for effective learning from both types of samples. We introduce a novel cross-architecture approach where 3D Convolutional Neural Networks (3D CNNs) and video transformers (VIT) are utilized to capture different aspects of action representations; hence we call it \textit{ActNetFormer}. The 3D CNNs excel at capturing spatial features and local dependencies in the temporal domain, while VIT excels at capturing long-range dependencies across frames. By integrating these complementary architectures within the ActNetFormer framework, our approach can effectively capture both local and global contextual information of an action. This comprehensive representation learning enables the model to achieve better performance in semi-supervised action recognition tasks by leveraging the strengths of each of these architectures. Experimental results on standard action recognition datasets demonstrate that our approach performs better than the existing methods, achieving state-of-the-art performance with only a fraction of labeled data. The official website of this work is available at: \texttt{\url{https://github.com/rana2149/ActNetFormer}}.

\end{abstract}
\section{Introduction}

The remarkable advancements in deep learning have revolutionized action recognition, particularly with the advent of supervised learning protocols. However, acquiring a substantial number of annotated videos remains a challenge in practice since it is time-consuming and expensive~\cite{zhu2020comprehensive,pareek2021survey}. Each day, video-sharing platforms like YouTube and Instagram witness millions of new video uploads. Leveraging this vast pool of unlabeled videos presents a significant opportunity for semi-supervised learning approaches, promising substantial benefits for advancing action recognition capabilities~\cite{shen2015evaluation,zhang2011boosted}.

A typical method for leveraging unlabeled data involves assigning pseudo-labels to them and effectively treating them as \textit{ground truth} during training \cite{Hu_2021_CVPR,Wang_2022_CVPR,sohn2020fixmatch}. Current methodologies typically involve training a model on annotated data and subsequently employing it to make predictions on unlabeled videos. When predictions exhibit high confidence levels, they are adopted as pseudo-labels for the respective videos, guiding further network training. However, the efficacy of this approach hugely depends on the quantity and accuracy of the pseudo-labels generated. Unfortunately, the inherent limitations in discriminating patterns from a scant amount of labeled data often result in subpar pseudo-labels, ultimately impeding the potential benefits gleaned from unlabeled data. 

\begin{figure}[t]
\centering
\begin{subfigure}[b]{0.48\textwidth}
  \centering
  \includegraphics[width=\textwidth]{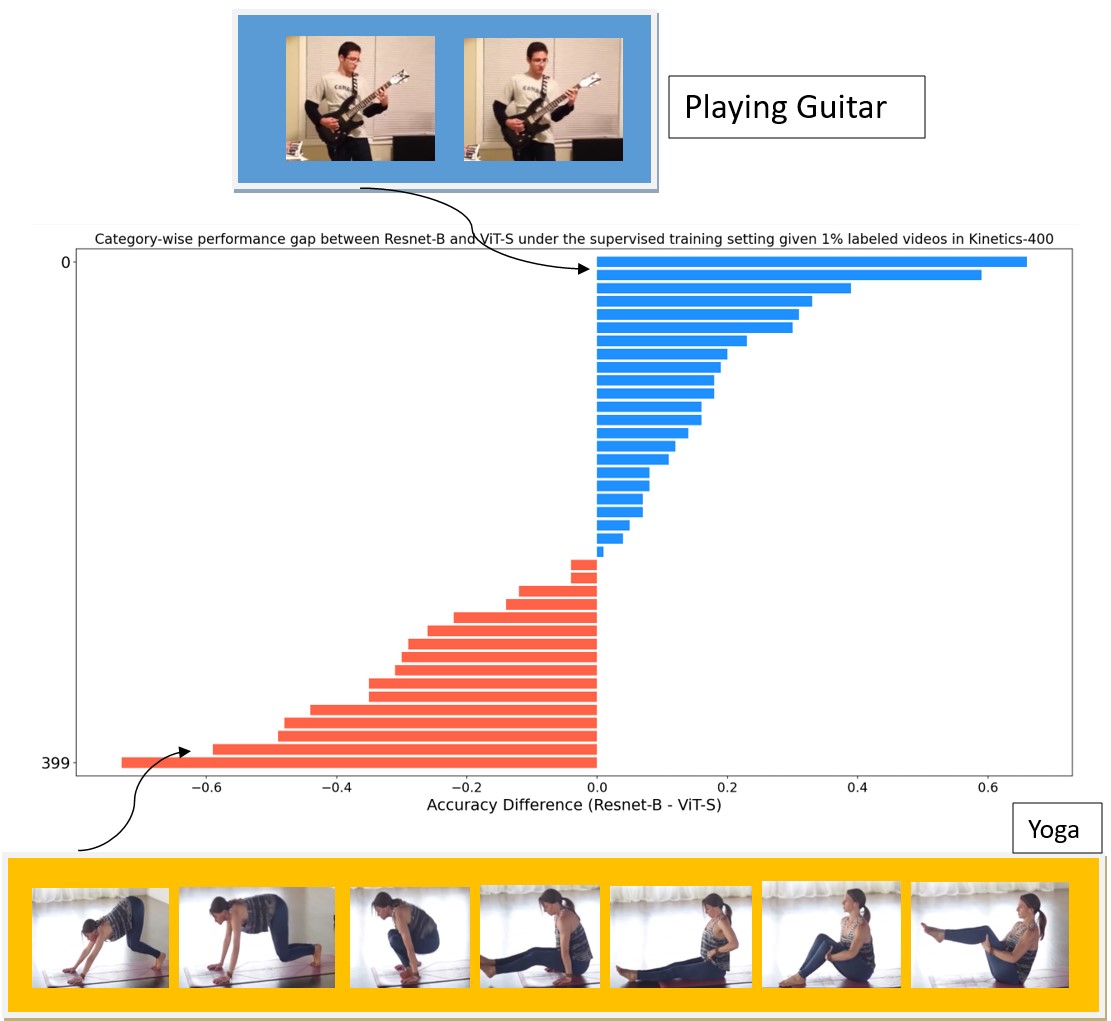}
  \caption{The difference in performance between ResNet-B and VIT-S, categorized by class, is evaluated under a supervised training scenario with only 1\% labeled videos in the Kinetics-400 dataset.}
  \label{fig:moti1}
\end{subfigure}
\hfill
\begin{subfigure}[b]{0.48\textwidth}
  \centering
  \includegraphics[width=\textwidth]{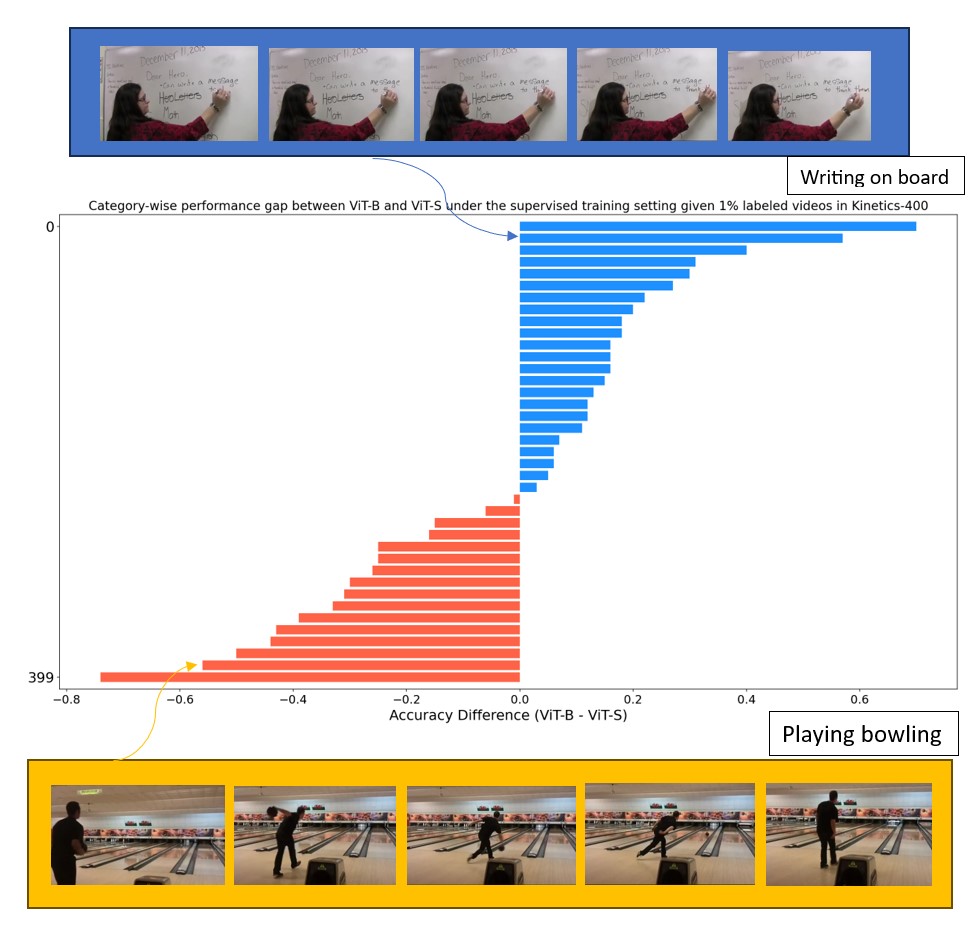}
  \caption{The difference in performance between VIT-B and VIT-S, categorized by class, is evaluated under a supervised training scenario with only 1\% labeled videos in the Kinetics-400 dataset.}
  \label{fig:moti2}
\end{subfigure}
\caption{Comparison of performance between different architectural models.}
\label{fig:moti}
\end{figure}

To enhance the utilization of unlabeled videos, our approach draws inspiration from recent studies, particularly from~\cite{xu2022cross}, which introduced an auxiliary model to provide complementary learning. We also introduce complementary learning but with notable advancements. Firstly, we introduce a cross-architecture strategy, leveraging both 3D CNNs and transformer models' strengths, unlike CMPL~\cite{xu2022cross}, which relies solely on 3D CNNs. This is because both 3D CNNs and video transformers (VIT) offer distinct advantages in action recognition. As shown in Fig.~\ref{fig:moti1}, videos for 
activities such as 
`playing the guitar'' from the Kinetics-400 dataset that demonstrate short-range temporal dependencies typically involve actions or events that occur over a relatively short duration and require capturing temporal context within a limited time-frame, and perform better with 3D CNNs. This is because 3D CNNs excel at capturing spatial features and local dependencies in the temporal domain due to their intrinsic property, which involves processing spatio-temporal information through convolutions.

On the other hand, transformer architectures, leveraging self-attention mechanisms, can naturally capture long-range dependencies by allowing each token to learn attention across the entire sequence. As shown in Fig.~\ref{fig:moti1} videos such as the "yoga'' class in the Kinetics-400 dataset, which demonstrate long-range temporal dependencies involving actions or events that unfold gradually over extended periods that require capturing temporal context over more extended periods, perform better in the transformer model. Such intrinsic property in transformers enables them to capture complex relationships and interactions between distant frames, leading to a more holistic understanding of the action context. This capability enables transformers to encode meaningful context information into video representations, facilitating a deeper understanding of the temporal dynamics and interactions within the video sequence.


Besides that, CMPL~\cite{xu2022cross} also suggests that smaller models excel at capturing temporal dynamics in action recognition. In comparison, larger models are more adept at learning spatial semantics to differentiate between various action instances. Motivated by this approach, we chose to leverage the advantages of a smaller transformer model, VIT-S, over its larger counterpart, VIT-B. As depicted in Fig.~\ref{fig:moti2} and further studied in Section S2 in the \textit{Supplementary Material}, a smaller model, despite its smaller capacity, does obtain significant improvements over a bigger model in certain classes. While VIT-B excels at capturing spatial semantics, it is essential to note that our primary model, 3D-ResNet50, already possesses these strong capabilities. The 3D convolutional nature of ResNet-50 makes it well-suited for extracting spatial features and local dependencies within the temporal domain. Therefore, the inclusion of VIT-S as an auxiliary model complements the strengths of our primary model by focusing on capturing temporal dynamics, which aligns with our primary objective of addressing action recognition in videos. This strategic combination allows our ActNetFormer framework to achieve a balanced representation learning, leveraging the spatial semantics captured by 3D-ResNet50 and the temporal dynamics captured by VIT-S. As demonstrated in our ablation study (Section~\ref{sec:analysis_pa}), this integration of VIT-S as an auxiliary model consistently leads to better results compared to adapting VIT-B. Hence, while VIT-B remains essential, its role is effectively supported by the capabilities of our primary model, thereby justifying our choice of prioritizing VIT-S within the ActNetFormer framework.

Furthermore, our method also incorporates video level contrastive learning, enabling the model to glean stronger representations at the spatio-temporal level. Hence, our cross-architecture pseudo-labeling approach is utilized to capture distinct aspects of action representation from both the 3D CNNs and transformer architectures, while cross-architecture contrastive learning aims explicitly to align the representations and discover mutual information in global high-level representations across these architectures. More experimental details about the cross-architecture strategy are included in Section S1.1 in the \textit{Supplementary Material}. 

The main contributions of this work is twofold and listed as follows: 
\begin{itemize}
  \item We propose a novel cross-architecture pseudo-labeling framework for semi-supervised action recognition in videos.  
  \item An architecture-level contrastive learning is developed to enhance the performance of the proposed approach for action recognition in videos. 

\end{itemize}

\begin{figure*}[t]
\centerline{\includegraphics[width=\linewidth]{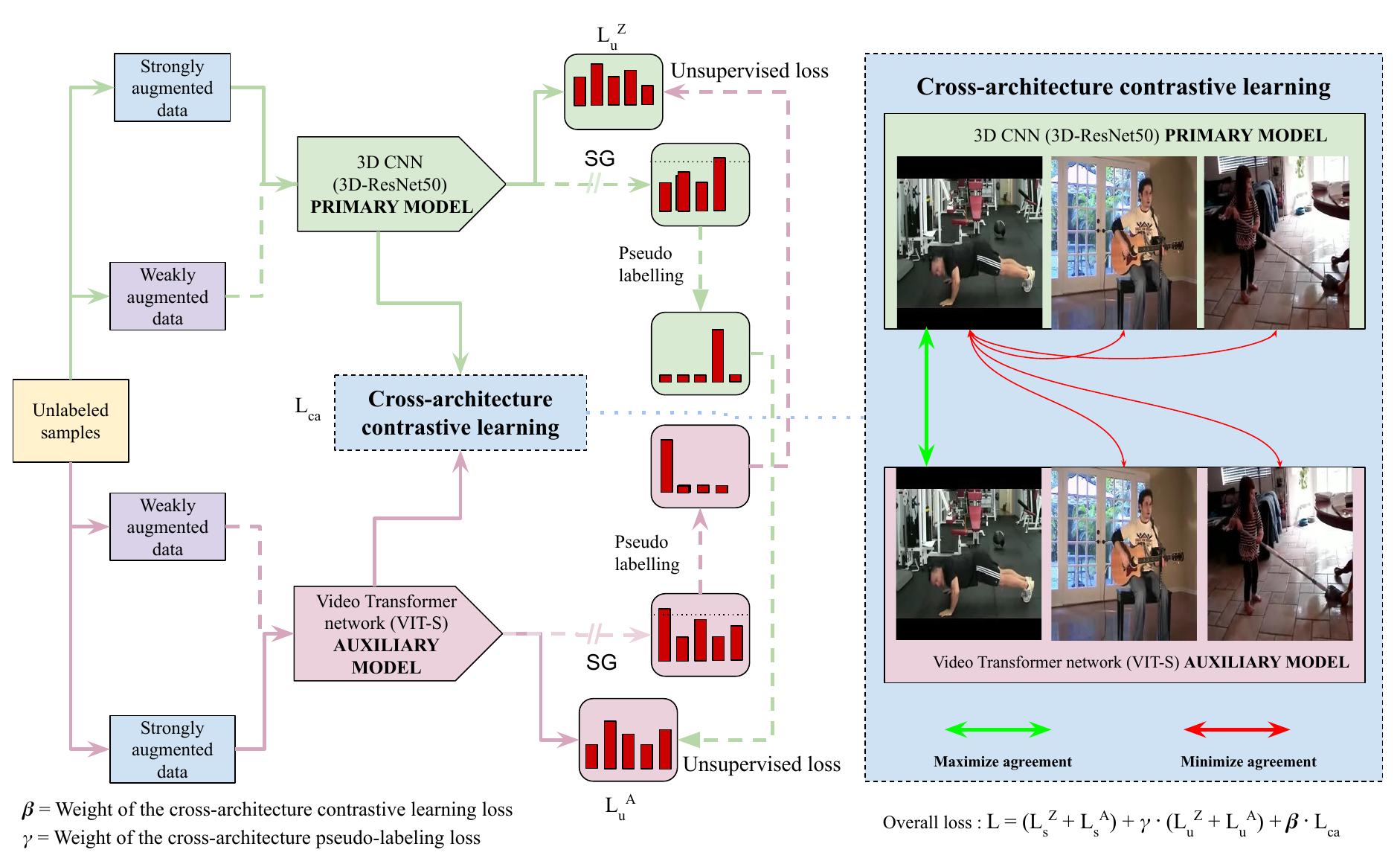}}
\caption{Architecture of the proposed framework.}
 \label{fig:framework}
\end{figure*}

\section{Related works}

\subsection{Action Recognition}
Action recognition has advanced significantly with deep learning architectures like CNNs, Recurrent Neural Networks (RNNs), Long Short-term Memory (LSTM), and Transformers. CNNs capture spatial information, while the RNNs captures temporal dependencies. Meanwhile, Transformers, known for NLP tasks, is excellent at capturing long-range dependencies. Varshney \etal~\cite{varshney2022deep} proposed a CNN model combining spatial and temporal information using different fusion schemes for human activities recognition video. Bilal \etal~\cite{bilal2022transfer} employ hybrid CNN-LSTM models and transfer learning for recognizing overlapping human actions in long videos. Vision Transformer (ViT)~\cite{dosovitskiy2020image} treats images as sequences of patches, achieving competitive performance on image classification tasks. Arnab \etal~\cite{arnab2021vivit} extend Transformers to video classification, while Bertasius \etal~\cite{bertasius2021space} introduce TimeSformer, a convolution-free approach to video classification built exclusively on self-attention over space and time convolution-free approach. TimeSformer achieves state-of-the-art (SOTA) results on action recognition benchmarks like Kinetics-400 and Kinetics-600, offering faster training and higher efficiency. Besides that, TimeSformer can also achieve good results even without pretraining. However, achieving these results may require more extensive data augmentation and longer training periods.


\subsection{Semi-supervised Learning for Video Action Recognition}

Action recognition in computer vision is vital across various applications, yet it often suffers from limited labeled data. Semi-supervised learning (SSL) methods provide a solution by utilizing both labeled and unlabeled data to enhance model performance~\cite{sohn2020fixmatch,xu2022cross}. These approaches exploit the abundance of unlabeled video data available online. Wu \etal~\cite{wu2023neighbor} proposed NCCL, a neighbor-guided consistent and contrastive learning (NCCL) method for semi-supervised video-based action recognition. Xu \etal~\cite{xu2022cross} introduced CMPL, employing cross-model predictions to generate pseudo-labels and improve model performance. Singh \etal~\cite{singh2021semi} leverage unsupervised videos played at different speeds to address limited labeled data. Xiao \etal~\cite{xiao2022learning} enhance semi-supervised video action recognition by incorporating temporal gradient information alongside RGB data. Jing \etal~\cite{jing2022tp} use pseudo-labels from CNN confidences and normalized probabilities to guide training, achieving impressive results with minimal labeled data. Gao \etal~\cite{gao2023danet} introduced an end-to-end semi-supervised Differentiated Auxiliary guided Network (DANet) for action recognition. Xiong \etal~\cite{Xiong_2021_ICCV} introduce multi-view pseudo-labeling, leveraging appearance and motion cues for improved SSL. Tong \etal~\cite{9904603} propose TACL, employing temporal action augmentation, action consistency learning, and action curriculum pseudo-labeling for enhanced SSL. These advancements demonstrate the potential of SSL techniques in boosting action recognition performance, especially in scenarios with limited labeled data. 

\subsection{Contrastive Learning in Action Recognition}

Contrastive learning has become a popular approach, especially in computer vision~\cite{le2020contrastive}. Unlike supervised methods, contrastive learning operates on unlabeled data, maximizing agreement between similar samples while minimizing it between dissimilar ones~\cite{shah2022max, tian2020makes}. It fosters a feature space where similar instances are clustered and dissimilar ones are separated. By optimizing a similarity metric using positive (similar) and negative (dissimilar) sample pairs, contrastive learning extracts meaningful features beneficial for tasks like classification and object detection. Its advantage lies in learning from vast unlabeled data, making it suitable for scenarios with limited labeled data~\cite{dave2022tclr, tian2022understanding}. Guo \etal~\cite{Guo_Liu_Chen_Liu_Wang_Ding_2022} propose AimCLR, a contrastive learning-based self-supervised action representation framework. They enhance positive sample diversity and minimize distribution divergence, achieving superior performance. 
The method in \cite{zheng2022few} also proposes a hierarchical matching model for few-shot action recognition, leveraging contrastive learning to enhance video similarity measurements across multiple levels. Rao \etal~\cite{rao2021augmented} introduce AS-CAL, a contrastive learning-based approach for action recognition using 3D skeleton data. It learns inherent action patterns across various transformations, facilitating effective representation of human actions.

\section{Method}

\subsection{Overview of our work}

The proposed ActNetFormer framework is illustrated in Fig.~\ref{fig:framework}. Our approach consists of two models, \ie, the primary model $Z(\cdot)$ and the auxiliary model $A(\cdot)$. These models process video inputs with varying frame rates, utilizing 3D-ResNet50 as the primary model and VIT-S as the auxiliary model by default. When presented with an unlabeled video, both models independently generate predictions on the data that are weakly augmented. The predicted outcomes are then utilized to generate a pseudo-label for the counterpart model, acting as guidance for the strongly augmented data. The ``SG" notation denotes the stop-gradient operation, and supervised losses from labeled data are not depicted in this figure. Additionally, we incorporate contrastive learning to maximize agreement between the outputs of the two architectures for the same video while minimizing the agreement for different videos. ActNetFormer leverages the strengths of both a 3D CNN and a transformer. Given an input video clip, each model produces a video representation separately. This encourages each model to focus on different features or patterns within the videos, leading to more comprehensive representations. By combining these complementary representations through contrastive learning, the framework can leverage a richer set of features for action recognition.

\subsection{Our proposed framework}

Given a labeled dataset $X$ containing $N_l$ videos, each paired with a corresponding label $(x_i, y_i)$, and an unlabeled dataset $U$ comprised of $N_u$ videos, ActNetFormer efficiently learns an action recognition model by utilizing both data that are labeled and unlabeled. Typically, the size of the unlabeled dataset $N_u$ is greater than that of the labeled dataset $N_l$. We provide a brief description of the pseudo-labeling method in Section~\ref{section.ppl}. Subsequently, we introduce the proposed ActNetFormer framework in Section~\ref{section.CAPL}. Then, we explain how contrastive learning works in ActNetFormer framework in Section~\ref{section.CL}. Subsequently, we delve into the implementation details of ActNetFormer in Section~\ref{section.Implel}.

\subsection{Preliminaries on Pseudo-Labeling}\label{section.ppl}

Pseudo-labeling is a widely employed approach in semi-supervised image recognition, aiming to leverage the model to generate artificial labels for data that are not labeled~\cite{sohn2020fixmatch,zhu2020comprehensive,zhang2021flexmatch}. The generated labels that surpass a predefined threshold are kept, enabling the associated unlabeled data to be utilized as extra samples for training. FixMatch~\cite{sohn2020fixmatch}, a recent SOTA approach, utilizes weakly augmented images for acquiring pseudo-labels, which are subsequently combined with strongly augmented versions to generate labeled samples. The extension of FixMatch to semi-supervised action recognition can be accomplished as follows:

\begin{equation}\label{equ.1}
L_u = \frac{1}{B_u} \sum_{i=1}^{B_u} 1\left(\max(q_i) \geq \tau\right)\mathcal{H}\left(\hat{y}_i, Z(G_{\text{s}}(u_i))\right),
\end{equation}

In the equation (\ref{equ.1}), $B_u$ denotes the batch size, $\tau$ is the threshold used to indicate if the prediction that is made is reliable or not, $1(\cdot)$ denotes the indicator function, $q_i = Z(G_{\text{w}}(u_i))$ represents the class distribution, and $\hat{y}_i = \arg\max(q_i)$ denotes the pseudo-label. $G_{\text{s}}(\cdot)$ and $G_{\text{w}}(\cdot)$ respectively denote the processes of strong and weak augmentation. $\mathcal{H}(\cdot, \cdot)$ represents the standard cross-entropy loss. $L_u$ represents the loss on the unlabeled data, while the loss on the labeled data is the cross-entropy loss typically used in action recognition.

\subsection{Cross-Architecture Pseudo-Labeling}\label{section.CAPL}

In Section~\ref{section.ppl}, we discussed the fundamental concept underlying recent semi-supervised learning methodologies, which revolves around generating high-quality pseudo-labels for unlabeled data. However, in scenarios where the number of labeled instances is constrained, a single model may lack the necessary discriminative power to assign pseudo-labels effectively to a large volume of unlabeled data~\cite{xu2022cross}. 
To address this challenge our approach (Cross-Architecture Pseudo-Labeling in ActNetFormer) adopts a novel strategy of employing two models with distinct architectures and tasking them with generating pseudo-labels for each other. This approach is influenced by the understanding that different models exhibit distinct strengths and biases. While 3D CNNs excel in capturing spatial features and local dependencies within the temporal domain, transformers are more adept at handling long-range dependencies within the temporal domain. This variation in architectural characteristics leads to the generation of complementary semantic representations.

As shown in Fig.~\ref{fig:framework}, we illustrate the ActNetFormer framework, which employs a cross-architecture setup. Specifically, we utilize the 3D-ResNet50 as the primary model \( Z(\cdot) \) and video transformer (VIT-S) as the auxiliary model \( A(\cdot) \). Both models undergo supervised training using labeled data while simultaneously providing pseudo-labels for data unlabeled to their counterparts. This method encourages the two architectures to understand complementary representations, ultimately enhancing overall efficacy.

\subsubsection{Training on labeled data.}
Training a model on labeled data involves a straightforward process. Given a set of labeled videos $\{(x_i, y_i)\}_{i=1}^{B_l}$, we define the supervised loss for both models as follows: 

\begin{equation}\label{equ.2}
L^{\mathrm{Z}}_{s} = \frac{1}{B_l} \sum_{i=1}^{B_l} \mathcal{H}(y_i, Z(G_{\text{n}}^Z(x_i)))
\end{equation}

\begin{equation}\label{equ.3}
L^{\mathrm{A}}_{s} = \frac{1}{B_l} \sum_{i=1}^{B_l} \mathcal{H}(y_i, A(G_{\text{n}}^A(x_i)))
\end{equation}

where $G_{\text{n}}(\cdot)$ denotes the conventional data augmentation method employed in~\cite{Wang_2018_CVPR,feichtenhofer2019slowfast}.

\subsubsection{Training on unlabeled data.}
When presented with an unlabeled video \(u_i\), the auxiliary model \(A(\cdot)\) generates predictions based on data that are weakly augmented \(u_i\) and produces category-wise probabilities denoted as \(q^{A}_i = A(G_{\text{w}}(u_i))\). If the maximum probability among these probabilities, \(\max(q^{A}_i)\), exceeds a predefined threshold \(\tau\), it is considered a reliable prediction. In such cases, we utilize \(q^{A}_i\) to infer the pseudo ground truth label \(\hat{y}^{A}_{i} = \arg \max(q^{A}_i)\) for the strongly augmented \(u_i\). This process allows the model \(Z(\cdot)\) to learn effectively.

\begin{equation}\label{equ.4}
L^{Z}_{u} = \frac{1}{B_u} \sum_{i=1}^{B_u} 1\left(\max(q^{A}_{i}) \geq \tau\right)\mathcal{H}(\hat{y}^{A}_{i}, Z(G_{\text{s}}(u_i)))
\end{equation}
where, $B_u$ represents the batch size, and $\mathcal{H}(\cdot, \cdot)$ denotes the cross-entropy loss.

Similar to the auxiliary model, the primary model will also produce a prediction \(q^{Z}_i = Z(G_{\text{w}}(u_i))\), which is then utilized to create a labeled pair $(\hat{y}^{Z}_{i}, G_{\text{s}}(u_i))$ for the auxiliary model:

\begin{equation}\label{equ.5}
L^{A}_{u} = \frac{1}{B_u} \sum_{i=1}^{B_u} 1\left(\max(q^{Z}_{i}) \geq \tau\right)\mathcal{H}(\hat{y}^{Z}_{i}, A(G_{\text{s}}(u_i)))
\end{equation}

\subsubsection{Contrastive learning.}\label{section.CL}

The goal is to train the primary and auxiliary models using limited supervision initially, which can effectively analyze a vast collection of unlabeled videos to enhance activity understanding. Our cross-architecture pseudo-labeling approach already leverages two different architectures to capture different aspects of action representations as mentioned in Section~\ref{section.CAPL}. Contrastive learning is incorporated to encourage the models further to extract complementary features from the input data, leading to more comprehensive representations of actions. 3D CNN and a Video Transformer process the input video clip differently and produce a unique representation of the video content. In other words, the features extracted by each architecture capture different aspects of the video, such as spatial and temporal information. This diversity in representations can be advantageous as it allows the model to learn from multiple perspectives, potentially leading to a more comprehensive understanding of the action sequences in the videos. Therefore, cross-architecture contrastive learning is employed to discover the mutual information that coexists between both the representation encoding generated by the 3D CNN and the video transformer model. It is worth noting that our framework uses weakly augmented samples from each architecture for cross-architecture contrastive learning, inspired by \cite{xiao2022learning}.

Consider a mini-batch with $B_u$ unlabeled videos. Here, $m(u^Z_i)$ represents the video clip processed by the primary model, while $m(u^A_i)$ represents the video clip processed by the auxiliary model. Therefore, $m$ can be interpreted as the function that generates representations of the input videos through the respective models.  These representations form the positive pair. For the rest of $B_u - 1$ videos, $m(u^Z_i)$ and $m(u^q_k)$ form negative pairs, where the representation of the $k$-th video can come from either of the architecture (i.e., $q \in \{Z, A\}$). Given that the negative pairs comprise various videos with distinct content, the representation of different videos within each architecture is pushed apart. This is facilitated by utilizing a contrastive loss ($L_{ca}$) adapted from~\cite{singh2021semi, pmlr-v119-chen20j}, as outlined below.

\begin{equation}\label{equ.6}
L_{ca}(u^Z_i, u^A_i) = -\log \frac{{h(m(u^Z_i), m(u^A_i))}}{{h(m(u^Z_i), m(u^A_i)) + \sum_{{k=1}\atop q\in\{Z,A\}}^{B}1_{\{k \neq i\}} h(m(u^Z_i), m(u^q_k))}}
\end{equation}
where, \( h(u, v) \) = $\exp \left( \frac{u^\top v}{\| u \|_2 \| v \|_2} /\tau \right) $ represents the exponential of the cosine similarity measure between vectors \( u \) and \( v \), where \( \tau \) denotes the temperature hyperparameter. The final contrastive loss is calculated for all positive pairs, \( (u^Z_i, u^A_i) \), where \( u^Z_i\) is the representation generated by the primary model and \( u^A_i\) is the representation generated by auxiliary model. The loss function is engineered to reduce the similarity, not just among different videos processed within individual architectures but also across both architectural models.

\subsubsection{Complete Training objective.}

To encapsulate, merging supervised losses derived from labeled data with unsupervised losses derived from unlabeled data, we present the entire objective function as:

\begin{equation}\label{equ.7}
L = (L^{\mathrm{Z}}_{s} + L^{\mathrm{A}}_{s}) + \gamma \cdot (L^{Z}_{u} + L^{A}_{u}) + \beta \cdot L_{\text{ca}}
\end{equation}
where, \( \gamma \) and \( \beta \) are weights of the cross-architecture loss and contrastive learning losses respectively.

\section{Implementation}\label{section.Implel}
\subsection{Auxiliary Model}
As mentioned in Section~\ref{section.CAPL}, the auxiliary model should possess distinct learning capabilities compared to the primary model in order to offer complementary representations. Hence, we utilize VIT-S, which is the smaller version of the bigger transformer model (VIT-B). Comprehensive ablation studies (in the next section) show the superiority of VIT-S \wrt the transformer model (VIT-B) and the smaller 3D CNN model (3D-ResNet18). Unless otherwise specified, we utilize 3D-ResNet50 as the primary and VIT-S as the auxiliary models, respectively. More details of these models are included in Section S3 in the \textit{Supplementary Material}.
\subsection{Spatial data augmentations}
We strictly adhere to the spatial data augmentations proposed in~\cite{feichtenhofer2019slowfast, Wang_2018_CVPR} for training, denoted as \( G_{\text{n}}(\cdot) \), on labeled data. For unlabeled data, random horizontal flipping, random scaling, and random cropping are employed as weak augmentations, denoted as \( G_{\text{w}}(\cdot) \). The input size of the video is standardized to \( 224 \times 224 \) pixels to ensure consistency during augmentation and subsequent processing by the models. We utilize techniques such as AutoAugment~\cite{Cubuk_2019_CVPR} or Dropout~\cite{bouthillier2016dropout} as strong augmentation, \( G_{\text{s}}(\cdot) \).

\subsection{Temporal data augmentations}

Our ActNetFormer framework incorporates variations in frame rates for temporal data augmentations inspired by prior research in~\cite{yang2020video,singh2021semi}. While the primary model operates at a lower frame rate, the auxiliary model is provided with a higher one. This variation in frame rates allows for exploring different speeds in video representations. Despite the differences in playback speeds, the videos maintain the same semantics, maximizing the similarity between their representations. This approach offers complementary benefits by leveraging both slower and faster frame rates between the primary and auxiliary models. Consequently, this contributes to improving the overall performance of our ActNetFormer framework in action recognition. Additional spatial and temporal augmentations analysis are provided in Section S1.2 in \textit{Supplementary Material}. 

\section{Experiments}

We assess the effectiveness of the proposed ActNetFormer framework on two widely used datasets, \ie, Kinetics-400~\cite{kay2017kinetics} and UCF-101~\cite{soomro2012ucf101}. We employ two standard settings for semi-supervised action recognition, \ie, 1\% and 10\% labeled data. Detailed ablation studies on the design choices of ActNetFormer are also conducted. Additionally, empirical analysis is provided in Section S2 in the \textit{Supplementary Material} to validate the motivations behind ActNetFormer. It is crucial to emphasize that all experiments are conducted using a single modality (RGB only) and assessed on the corresponding validation sets unless stated otherwise.

\subsection{Dataset}

The Kinetics-400 dataset~\cite{kay2017kinetics} comprises a vast collection of human action videos, encompassing around 245,000 training samples and 20,000 validation samples across 400 distinct action categories. Following established methodologies like MvPL~\cite{Xiong_2021_ICCV} and CMPL~\cite{xu2022cross}, we adopt a labeling rate of 1\% or 10\%, selecting 6 or 60 labeled training videos per category. Additionally, the UCF-101 dataset~\cite{soomro2012ucf101} offers 13,320 video samples spread across 101 categories. We also sample 1 or 10 samples in each category as the labeled set following CMPL~\cite{xu2022cross}.

\subsection{Baseline}

For our primary model, we utilize the 3D-ResNet50 from~\cite{feichtenhofer2019slowfast}. 
We employ the ViT~\cite{dosovitskiy2020image} extended with the video TimeSformer~\cite{bertasius2021space} as the auxiliary model in our ActNetFormer approach. While most hyperparameters remain consistent with the baseline, we utilize the divided space-time attention mechanism, as mentioned in TimeSformer~\cite{bertasius2021space}. However, only the big transformer model (VIT-B) is offered in TimeSformer, hence we adopt the smaller transformer model (VIT-S) inspired by DeiT-S~\cite{pmlr-v139-touvron21a} with the dimensions of 384 and 6 heads. More details on the structure of primary and auxiliary models are included in Section S3 in the \textit{Supplementary Material}. 

\subsection{Training and inference}
During training, we utilize a stochastic gradient descent (SGD) optimizer with a momentum of 0.9 and a weight decay of 0.001. The confidence score threshold $\tau$, is set to 0.8. Parameters $\gamma$ and $\beta$  are both set to 2. Based on insights from the ablation study in Section~\ref{sec:aoh}, we employ a batch ratio of 1:5 for labeled to unlabeled data, ensuring a balanced and effective training process. A total of 250 training epochs are used. During testing, consistent with the inference method employed in MvPL~\cite{Xiong_2021_ICCV} and CMPL~\cite{xu2022cross}, we uniformly sample five clips from each video and generate three distinct crops to achieve a resolution of 224 $\times$ 224, covering various spatial areas within the clips. The final prediction is obtained by averaging the softmax probabilities of these 5 $\times$ 3 predictions. While both the primary and auxiliary models are optimized jointly during training, only the primary model is utilized for inference, thereby incurring no additional inference cost. It is noteworthy that our ActNetFormer approach does not rely on pre-training or pre-trained weights, setting it apart from other methods and underscoring its uniqueness in the field of action recognition in videos. 

\section{Results}

\begin{table*}[t]
\centering
\resizebox{\textwidth}{!}{%
\begin{tabular}{l|l|l|l|l|l|l|l}
\toprule[0.5mm]
Method & Backbone & Input & Epoch & \multicolumn{2}{l|}{UCF-101} & \multicolumn{2}{l}{Kinetics-400} \\ \cline{5-8} 
 &  &  &  & 1\% & 10\% & 1\% & 10\% \\ \midrule[0.25mm]
FixMatch (NeurIPS 2020) \cite{sohn2020fixmatch} & 3D-ResNet50 & V & 200 & 14.8 & 49.8 & 8.6 & 46.9 \\
FixMatch (NeurIPS 2020)  \cite{sohn2020fixmatch} & SlowFast-R50 & V & 200 & 16.1 & 55.1 & 10.1 & 49.4 \\
TCL (CVPR 2021) \cite{singh2021semi} & TSM-ResNet-18 & V & 400 & - & - & 8.5 & - \\
MvPL (ICCV 2022) \cite{Xiong_2021_ICCV} & 3D-ResNet50 & V+F+G & 600 & 22.8 & \textcolor{blue}{80.5} & 17.0 & 58.2 \\\textcolor{gray}{TACL (IEEE TCSVT 2022)\cite{9904603}} & \textcolor{gray}{3D-ResNet50} & \textcolor{gray}{V} & 
\textcolor{gray}{200} & \textcolor{gray}{-} & \textcolor{gray}{55.6} &\textcolor{gray}{-} & \textcolor{gray}{-} \\

LTG (CVPR 2022) \cite{xiao2022learning} & 3D-ResNet18 & V+G & 180 & - & 62.4 & 9.8 & 43.8 \\
CMPL (CVPR 2022) \cite{xu2022cross} & 3D-ResNet50 & V & 200 & \textcolor{blue}{25.1} & 79.1 & \textcolor{blue}{17.6} & \textcolor{blue}{58.4} \\
NCCL (IEEE TIP 2023) \cite{wu2023neighbor} & TSM-ResNet-18 & V+G & 400 & 21.6 & - & 12.2 & 43.8 \\
DANet (Elsevier NN 2023) \cite{gao2023danet} & DANet & V & 600 & - & 64.6 & - & - \\
\Xhline{1pt} 
\textbf{ActNetFormer (Ours)} & 3D-ResNet50 & V & 250 & 26.1 & 80.0 & 18.3 & 59.2 \\ 
\textbf{ActNetFormer (Ours) with Contrastive learning} & 3D-ResNet50 & V & 250 & \textcolor{red}{27.6} & \textcolor{red}{80.6} & \textcolor{red}{19.1} & \textcolor{red}{59.8} \\ 
\bottomrule[0.5mm]

\end{tabular}%
}

\caption{Comparison of results with SOTA approaches on UCF-101 and Kinetics-400. The best-performing results are highlighted in \textcolor{red}{red}, while the second-best results are highlighted in \textcolor{blue}{blue}. Methods utilizing pre-trained ImageNet weights are displayed in \textcolor{gray}{grey}.}
\label{tab:comparison}
\end{table*}

The backbone column in Table~\ref{tab:comparison} denotes the primary model used in the respective methods. We present the top-1 accuracy as our chosen evaluation metric. The ``Input" category indicates the data format utilized during training, with ``V" representing raw RGB video, ``F" denoting optical flow, and ``G" indicating temporal gradient. ActNetFormer consistently performs better than various SOTA methods, including FixMatch, TCL, MvPL, TACL, CMPL, NCCL, DANet, and LTG, across both datasets and labeling rates. The inclusion of contrastive learning in our approach demonstrates an improved performance by a significant percentage, specifically at the 1\% labeled data setting. We observe a percentage increase of approximately 4.60\% for the UCF-101 and 4.37\% for the Kinetics-400 dataset. This enhancement underscores the effectiveness of incorporating contrastive learning, resulting in more robust representations. ActNetFormer outperforms FixMatch by a large margin due to its novel cross-architecture strategy, which leverages the strengths of both 3D CNN and VIT models, whereas FixMatch relies solely on its own architecture for label generation, potentially limiting its adaptability. Our approach shares similarities with the CMPL approach. However, it surpasses CMPL in several vital aspects. Firstly, our approach incorporates video-level contrastive learning, which enables the model to learn more robust representations at the video level. This enhanced representation leads to better performance in action recognition. Additionally, our approach leverages a cross-architecture strategy, combining the strengths of both 3D CNN and VIT models. In contrast,  CMPL leverages a cross-model strategy which utilizes the strength of 3D CNN alone. By integrating spatial feature extraction capabilities from CNNs with the attention mechanisms of transformers, our approach achieves a more comprehensive understanding of both spatial and temporal aspects of video data. Besides that, our approach achieves a performance of 80.0\% in the 10\% UCF-101 dataset, while incorporating contrastive learning boosts our performance to 80.6\%, bringing it closer to the 80.5\% achieved by MvPL. Notably, our approach relies solely on one modality, whereas MvPL exploits three modalities. Despite this discrepancy in input modalities, our approach demonstrates comparable performance, indicating its efficiency in leveraging single-modality information for video understanding tasks. This suggests that our approach may offer a more streamlined solution than MvPL, which relies on multiple modalities to achieve similar performance levels.

\section{Ablation Studies}

We thoroughly examine the proposed ActNetFormer method through several ablation studies. We present the experimental outcomes of various configurations of hyperparameters. We then analyze different combinations of the primary and auxiliary models. In all the ablation studies, it is crucial to highlight that experiments conducted with the UCF-101 dataset utilize 1\% of the labeled data, while those conducted with the Kinetics-400 dataset also employ 1\% of the labeled data.

\subsection{Analysis of hyperparameters}
\label{sec:aoh}

\begin{figure}[t]
\centerline{\includegraphics[scale=.5]{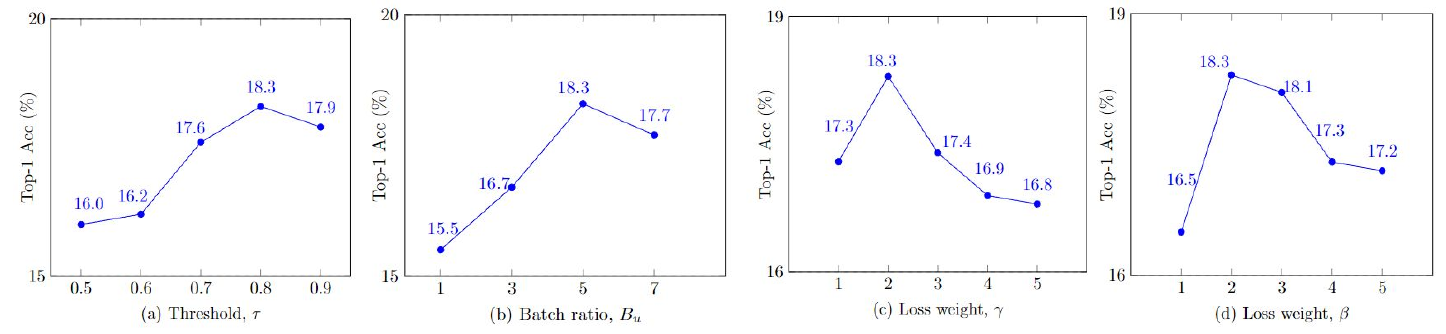}}
\caption{Analysis of different hyperparameters which includes Threshold $\tau$, Batch ratio $B_u$, Loss weight $\gamma$, Loss weight $\beta$.}
    \label{fig:main}
\end{figure}

Here, we investigate the impact of various hyperparameters. Experiments are conducted under the 1\% setting of the Kinetics-400 dataset. Initially, we examine the influence of different threshold values of $\tau$. As illustrated in Fig.~\ref{fig:main} (a), the results indicate that a threshold of ($\tau = 0.8$) achieved the highest accuracy, suggesting that the quality of the threshold is crucial. Additionally, setting the threshold too high, as in the case of ($\tau = 0.9$), may lead to sub-optimal performance, as evidenced by the lower accuracy compared to ($\tau = 0.8$). When the threshold is set too high, there is a risk that only a limited number of unlabeled samples are selected for inclusion. This occurs because the threshold acts as a criterion for determining which samples are considered confidently predicted by the model and thus eligible for inclusion in the training process. Therefore, if the threshold is excessively high, fewer unlabeled samples may meet this criterion, leading to under-utilization of unlabeled data and potentially compromising model performance. Hence, we utilize 0.8 as the threshold for all the experiments in this study. 

Next, we evaluate the impact of the ratio between labeled and unlabeled samples in a mini-batch on the final outcome. Specifically, we fix the number of labeled samples $B_l$ at 1 and randomly sample $B_u$ unlabeled samples to form a mini-batch, where $B_u$ varies from \{1, 3, 5, 7\}. The outcomes are depicted in Fig.~\ref{fig:main} (b), indicating that the model performs best when $B_u = 5$. Lastly, we explore the selection of the loss weights $\gamma$ and $\beta$, as shown in Fig.~\ref{fig:main} (c) and Fig.~\ref{fig:main} (d) for the cross-architecture loss and contrastive learning loss, respectively. We find that the optimum value of $\gamma$ and $\beta$ are 2. Hence, we utilize $\gamma$ = 2 and $\beta$ = 2 for all the experiments. 

\subsection{Analysis of different combination of primary and auxiliary models used}
\label{sec:analysis_pa}

\begin{table*}[t]
\centering
\begin{tabular}{l|c|c|c}
\toprule[0.5mm]
\textbf{Primary Model} & \textbf{Auxiliary Model} & \textbf{UCF-101 (1\%)} & \textbf{Kinetics-400 (1\%)} \\ \midrule[0.25mm]
VIT-B & VIT-S & 19.2 & 13.1 \\ 
VIT-B & ResNet-B & 20.9 & 13.9 \\ 
VIT-B & ResNet-S & 21.1 & 14.6 \\ 
ResNet-B & VIT-B & 23.7 & 16.9 \\ 
ResNet-B & ResNet-S & 25.1 & 17.6 \\ 
\textbf{ResNet-B} & \textbf{VIT-S} & \textbf{26.1} & \textbf{18.3} \\
\bottomrule[0.5mm]
\end{tabular}
\caption{Comparison of performance between primary and auxiliary models on UCF-101 (1\%) and Kinetics-400 (1\%) datasets.}
\label{tab:model_comparison}
\end{table*}

``ResNet-B" explicitly denotes the 3D-ResNet50 model, while ``ResNet-S" refers to the 3D-ResNet18 model. Correspondingly, ``VIT-S" represents the smaller variant of the video transformer model, while ``VIT-B" indicates the larger variant. Please keep these specific references in mind for clarity in our discussions. Before delving into the comparisons, it is important to note that we have critically analyzed why our approach (ResNet-B and VIT-S) outperforms other combinations. The comparison between ResNet-B and VIT-S versus alternative combinations is illustrated in Table~\ref{tab:model_comparison}, and the analysis is detailed below.

The comparison between ResNet-B and VIT-S versus alternative combinations reveals detailed insights. ResNet-B and VIT-S, demonstrate the significance of cross-architecture approaches in video recognition tasks. Significant performance enhancements are achieved by leveraging ResNet-B’s spatial feature extraction and VIT-S’s temporal understanding. Additionally, VIT-S’s superiority as an auxiliary model highlights the effectiveness of smaller models, particularly in the temporal domain, due to its smaller parameter count and better suitability for scenarios with limited data. When VIT-B is employed as the primary model among the first three combinations, its best performance is achieved when paired with ResNet-S. This outcome validates our motivation for employing a cross-architecture strategy and demonstrates the efficacy of using smaller models as auxiliary components. The complementary nature and the efficacy in the temporal domain of smaller models enhance the overall performance. Overall, this analysis emphasizes the pivotal role of the cross-architecture approach and the utilization of smaller models in improving video recognition performance, aligning with the motivation of our study. Further ablations are provided in Section S1 of the \textit{Supplementary Material}.

\section{Conclusion}

In conclusion, our proposed approach, ActNetFormer, combines cross-architecture pseudo-labeling with contrastive learning to offer a robust solution for semi-supervised video action recognition. By leveraging both labeled and unlabeled data, ActNetFormer effectively learns action representations by merging pseudo-labeling and contrastive learning techniques. This novel approach integrates 3D CNN and VIT to comprehensively capture spatial and temporal aspects of action representations. Additionally, cross-architecture contrastive learning is employed to explore mutual information between the encoding generated by 3D CNN and VIT. This strategy enhances the model's ability to learn from diverse perspectives, resulting in superior performance. The success of ActNetFormer underscores the effectiveness of leveraging diverse architectures and semi-supervised learning paradigms to advance action recognition in real-world scenarios.

\bibliographystyle{splncs04}
\bibliography{bibliography}

\begin{thebibliography}{10}
\providecommand{\url}[1]{\texttt{#1}}
\providecommand{\urlprefix}{URL }
\providecommand{\doi}[1]{https://doi.org/#1}

\bibitem{arnab2021vivit}
Arnab, A., Deghani, M., Heigold, G., Sun, C., Lu{\v{c}}i{\'c}, M., Scmid, C.: Vivit: A video vision transformer. In: Proceedings of the IEEE/CVF international conference on computer vision. pp. 6936--6946 (2021)

\bibitem{bertasius2021space}
Bertasius, G., Wang, H., Torresani, L.: Is space - time attention all you need for video understanding? In: ICML. vol.~3, p.~4 (2021)

\bibitem{bilal2022transfer}
Bilal, M., Maqsood, M., Yasmin, S., Hasan, N.U., Rho, S.: A transfer learning-based efficient spatiotemporal human action recognition framework for long and overlapping action classes. The Journal of Supercomputing  \textbf{78}(2),  2873--2908 (2022)

\bibitem{bouthillier2016dropout}
Bouthillier, X., Konda, K., Vincent, P., Memisevic, R.: Dropout as data augmentation (2016)

\bibitem{pmlr-v119-chen20j}
Chen, T., Kornblith, S., Norouzi, M., Hinton, G.: A simple framework for contrastive learning of visual representations. In: III, H.D., Singh, A. (eds.) Proceedings of the 37th International Conference on Machine Learning. Proceedings of Machine Learning Research, vol.~119, pp. 1597--1607. PMLR (13--18 Jul 2020), \url{https://proceedings.mlr.press/v119/chen20j.html}

\bibitem{Cubuk_2019_CVPR}
Cubuk, E.D., Zoph, B., Mane, D., Vasudevan, V., Le, Q.V.: Autoaugment: Learning augmentation strategies from data. In: Proceedings of the IEEE/CVF Conference on Computer Vision and Pattern Recognition (CVPR) (June 2019)

\bibitem{dave2022tclr}
Dave, I., Gupta, R., Rizve, M.N., Shah, M.: Tclr: Temporal contrastive learning for video representation. Computer Vision and Image Understanding  \textbf{219},  103406 (2022)

\bibitem{dosovitskiy2020image}
Dosovitskiy, A., Beyer, L., Kolesnikov, A., Weissenborn, D., Zhai, X., Unterthiner, T., Dehghani, M., Minderer, M., Heigold, G., Gelly, S., et~al.: An image is worth 16x16 words: Transformers for image recognition at scale. arXiv preprint arXiv:2010.11929  (2020)

\bibitem{feichtenhofer2019slowfast}
Feichtenhofer, C., Fan, H., Malik, J., He, K.: Slowfast networks for video recognition. In: Proceedings of the IEEE/CVF international conference on computer vision. pp. 6202--6211 (2019)

\bibitem{gao2023danet}
Gao, G., Liu, Z., Zhang, G., Li, J., Qin, A.K.: Danet: Semi-supervised differentiated auxiliaries guided network for video action recognition. Neural Networks  \textbf{158},  121--131 (2023)

\bibitem{Guo_Liu_Chen_Liu_Wang_Ding_2022}
Guo, T., Liu, H., Chen, Z., Liu, M., Wang, T., Ding, R.: Contrastive learning from extremely augmented skeleton sequences for self-supervised action recognition. Proceedings of the AAAI Conference on Artificial Intelligence  \textbf{36}(1),  762--770 (Jun 2022). \doi{10.1609/aaai.v36i1.19957}, \url{https://ojs.aaai.org/index.php/AAAI/article/view/19957}

\bibitem{Hu_2021_CVPR}
Hu, Z., Yang, Z., Hu, X., Nevatie, R.: Simple: Similar pseudo label exploittation for semi - supervised clasiffication. In: Procedings of the IEEE/CVF Conference on Computer Vision and Pattern Recognition (CVPR). pp. 15099--15107 (June 2021)

\bibitem{jing2022tp}
Jing, Y., Wang, F.: Tp-vit: A two-pathway vision transformer for video action recognition. In: ICASSP 2022-2022 IEEE International Conference on Acoustics, Speech and Signal Processing (ICASSP). pp. 2185--2189. IEEE (2022)

\bibitem{kay2017kinetics}
Kay, W., Carreira, J., Simonyan, K., Zhang, B., Hillier, C., Vijayanarasimhan, S., Viola, F., Green, T., Back, T., Natsev, P., et~al.: The kinetics human action video dataset. arXiv preprint arXiv:1705.06950  (2017)

\bibitem{le2020contrastive}
Le-Khac, P.H., Healy, G., Smeaton, A.F.: Contrastive representation learning: A framework and review. Ieee Access  \textbf{8},  193907--193934 (2020)

\bibitem{pareek2021survey}
Pareek, P., Thakkar, A.: A survey on video-based human action recognition: recent updates, datasets, challenges, and applications. Artificial Intelligence Review  \textbf{54},  2259--2322 (2021)

\bibitem{rao2021augmented}
Rao, H., Xu, S., Hu, X., Cheng, J., Hu, B.: Augmented skeleton based contrastive action learning with momentum lstm for unsupervised action recognition. Information Sciences  \textbf{569},  90--109 (2021)

\bibitem{shah2022max}
Shah, A., Sra, S., Chellappa, R., Cherian, A.: Max-margin contrastive learning. In: Proceedings of the AAAI Conference on Artificial Intelligence. vol.~36, pp. 8220--8230 (2022)

\bibitem{shen2015evaluation}
Shen, H., Yan, Y., Xu, S., Ballas, N., Chen, W.: Evaluation of semi-supervised learning method on action recognition. Multimedia Tools and Applications  \textbf{74},  523--542 (2015)

\bibitem{singh2021semi}
Singh, A., Chakaborty, O., Varshney, A., Panda, R., Feris, R., Senko, K., Das, A.: Semi - supervised action recognition with temporal contrastive learning. In: Proceedings of the IEEE/CVF Conference on Computer Vision and Pattern Recognition. pp. 10389--10399 (2021)

\bibitem{sohn2020fixmatch}
Sohn, K., Berthelot, D., Carlini, N., Zhang, Z., Zhang, H., Raffel, C.A., Cubuk, E.D., Kurakin, A., Li, C.L.: Fixmatch: Simplifying semi-supervised learning with consistency and confidence. Advances in neural information processing systems  \textbf{33},  596--608 (2020)

\bibitem{soomro2012ucf101}
Soomro, K., Zamir, A.R., Shah, M.: Ucf101: A dataset of 101 human actions classes from videos in the wild. arXiv preprint arXiv:1212.0402  (2012)

\bibitem{tian2020makes}
Tian, Y., Sun, C., Poole, B., Krishnan, D., Schmmid, C., Isola, P.: What makes for good views for contrastive learning? Advances in neural information processing systems  \textbf{33},  6837--6839 (2020)

\bibitem{tian2022understanding}
Tian, Y.: Understanding deep contrastive learning via coordinate-wise optimization. Advances in Neural Information Processing Systems  \textbf{35},  19511--19522 (2022)

\bibitem{9904603}
Tong, A., Tang, C., Weng, W.: Semi - supervised action recognition from temporal augmentation using curriculum learning. IEEE Transactions on Circuits and Systems for Video Technology  \textbf{33}(3),  1301--1312 (2023). \doi{10.1109/TCSVT.2022.3310271}

\bibitem{pmlr-v139-touvron21a}
Touvron, H., Cord, M., Douze, M., Massa, F., Sablayrolles, A., Jegou, H.: Training data-efficient image transformers \& distillation through attention. In: Meila, M., Zhang, T. (eds.) Proceedings of the 38th International Conference on Machine Learning. Proceedings of Machine Learning Research, vol.~139, pp. 10347--10357. PMLR (18--24 Jul 2021), \url{https://proceedings.mlr.press/v139/touvron21a.html}

\bibitem{varshney2022deep}
Varshney, N., Bakariya, B.: Deep convolutional neural model for human activities recognition in a sequence of video by combining multiple cnn streams. Multimedia Tools and Applications  \textbf{81}(29),  42117--42129 (2022)

\bibitem{Wang_2018_CVPR}
Wang, X., Girshick, R., Gupta, A., He, K.: Non-local neural networks. In: Proceedings of the IEEE Conference on Computer Vision and Pattern Recognition (CVPR) (June 2018)

\bibitem{Wang_2022_CVPR}
Wang, Y., Wang, H., Shen, Y., Fei, J., Li, W., Jin, G., Wu, L., Zhao, R., Le, X.: Semi-supervised semantic segmentation using unreliable pseudo-labels. In: Proceedings of the IEEE/CVF Conference on Computer Vision and Pattern Recognition (CVPR). pp. 4248--4257 (June 2022)

\bibitem{wu2023neighbor}
Wu, J., Sun, W., Gan, T., Ding, N., Jiang, F., Shen, J., Nie, L.: Neighbor-guided consistent and contrastive learning for semi-supervised action recognition. IEEE Transactions on Image Processing  (2023)

\bibitem{xiao2022learning}
Xiao, J., Jing, L., Zhang, L., He, J., She, Q., Zho, Z., Yuile, A., Li, Y.: Learning from temporal gradient for semi - supervised action recognition. In: Proceedings of the IEEE/CVF Conference on Computer Vision and Pattern Recognition. pp. 3252--3262 (2022)

\bibitem{Xiong_2021_ICCV}
Xiong, B., Fan, H., Grauman, K., Feichtenhofer, C.: Multiview pseudo-labeling for semi-supervised learning from video. In: Proceedings of the IEEE/CVF International Conference on Computer Vision (ICCV). pp. 7209--7219 (October 2021)

\bibitem{xu2022cross}
Xu, Y., Wei, F., Sun, X., Yang, C., Shen, Y., Dai, B., Zhou, B., Lin, S.: Cross-model pseudo-labeling for semi-supervised action recognition. In: Proceedings of the IEEE/CVF Conference on Computer Vision and Pattern Recognition. pp. 2959--2968 (2022)

\bibitem{yang2020video}
Yang, C., Xu, Y., Dai, B., Zho, B.: Video representation learning with visual temporal consistency. arXiv preprint arXiv:2006.15599  (2020)

\bibitem{zhang2021flexmatch}
Zhang, B., Wang, Y., Hou, W., Wu, H., Wang, J., Okumura, M., Shinozaki, T.: Flexmatch: Boosting semi-supervised learning with curriculum pseudo labeling. Advances in Neural Information Processing Systems  \textbf{34},  18408--18419 (2021)

\bibitem{zhang2011boosted}
Zhang, T., Liu, S., Xu, C., Lu, H.: Boosted multi-class semi-supervised learning for human action recognition. Pattern recognition  \textbf{44}(10-11),  2334--2342 (2011)

\bibitem{zheng2022few}
Zheng, S., Chen, S., Jin, Q.: Few-shot action recognition with hierarchical matching and contrastive learning. In: European Conference on Computer Vision. pp. 297--313. Springer (2022)

\bibitem{zhu2020comprehensive}
Zhu, Y., Li, X., Liu, C., Zolfaghari, M., Xiong, Y., Wu, C., Zhang, Z., Tighe, J., Manmatha, R., Li, M.: A comprehensive study of deep video action recognition. arXiv preprint arXiv:2012.06567  (2020)

\end{thebibliography}


\begin{thebibliography}{1}
\providecommand{\url}[1]{\texttt{#1}}
\providecommand{\urlprefix}{URL }
\providecommand{\doi}[1]{https://doi.org/#1}

\bibitem{bouthillier2016dropout}
Bouthillier, X., Konda, K., Vincent, P., Memisevic, R.: Dropout as data augmentation (2016)

\bibitem{Cubuk_2019_CVPR}
Cubuk, E.D., Zoph, B., Mane, D., Vasudevan, V., Le, Q.V.: Autoaugment: Learning augmentation strategies from data. In: Proceedings of the IEEE/CVF Conference on Computer Vision and Pattern Recognition (CVPR) (June 2019)

\bibitem{Feichtenhofer_2019_ICCV}
Feichtenhofer, C., Fan, H., Malik, J., He, K.: Slowfast networks for video recognition. In: Proceedings of the IEEE/CVF International Conference on Computer Vision (ICCV) (October 2019)

\bibitem{singh2021semi}
Singh, A., Chakaborty, O., Varshney, A., Panda, R., Feris, R., Senko, K., Das, A.: Semi - supervised action recognition with temporal contrastive learning. In: Proceedings of the IEEE/CVF Conference on Computer Vision and Pattern Recognition. pp. 10389--10399 (2021)

\bibitem{sohn2020fixmatch}
Sohn, K., Berthelot, D., Carlini, N., Zhang, Z., Zhang, H., Raffel, C.A., Cubuk, E.D., Kurakin, A., Li, C.L.: Fixmatch: Simplifying semi-supervised learning with consistency and confidence. Advances in neural information processing systems  \textbf{33},  596--608 (2020)

\bibitem{touvron2021training}
Touvron, H., Cord, M., Douze, M., Massa, F., Sablayroles, A., J{\'e}gou, H.: Training data-efficient image transformers \& distillation through attention. In: International conference on machine learning. pp. 11347--11357. PMLR (2021)

\bibitem{yang2020video}
Yang, C., Xu, Y., Dai, B., Zho, B.: Video representation learning with visual temporal consistency. arXiv preprint arXiv:2006.15599  (2020)

\end{thebibliography}


%
%

\end{document}


\title{ActNetFormer: Transformer-ResNet Hybrid Pipeline for
Semi-Supervised Action Recognition in Videos (Supplementary Material) \thanks{This research is supported by the Global Research Excellence Scholarship, Monash University, Malaysia. This research is also supported, in part, by the Global Excellence and Mobility Scholarship (GEMS), Monash University, Malaysia \& Australia.}}
\titlerunning{Transformer-ResNet Hybrid Pipeline for
Semi-Supervised Action Recognition}
%
\author{Sharana Dharshikgan Suresh Dass\inst{1} \and
Hrishav Bakul Barua\inst{1,2} \and
Ganesh Krishnasamy\inst{1}
Raveendran Paramesran\inst{1} \and
Raphaël C.-W. Phan\inst{1}
}
\authorrunning{Dass et al.}
%
\institute{School of Information Technology, Monash University, Malaysia \and
Robotics and Autonomous Systems Group, TCS Research, India
\email{}\\
\url{} 
\email{\{sharana.sureshdass, hrishav.barua, ganesh.krishnasamy,\\raveendran.paramesran, raphael.phan\}@monash.edu}}
\maketitle              
\section{Additional Ablation Studies}

\subsection{Analysis on performance of different architecture  on UCF-101 (1\%) and Kinetics-400 (1\%) datasets under fully supervised settings}

\begin{table}[htbp]
    \centering
    \label{tab:performance_comparison}
    \begin{tabular}{l|c|c|c}
    \toprule[0.5mm]
    \textbf{Method} & \textbf{Backbone} & \textbf{UCF-101 (1\%)} & \textbf{Kinetics-400 (1\%)} \\ \midrule[0.25mm]
    Fully supervised & Resnet-B & 6.12 & 4.4 \\
    Fully supervised & Resnet-S & 4.07 & 2.23 \\
    Fully supervised & VIT-B & 5.01 & 2.51 \\
    Fully supervised & VIT-S & 5.47 & 2.85 \\ 
    \bottomrule[0.5mm]
    \end{tabular}
      \caption{Performance comparison on UCF-101 (1\%) and Kinetics-400 (1\%) datasets with different backbones under fully supervised settings.}
      \label{tab:model_comparison_label1}
\end{table}

``ResNet-B" explicitly denotes the 3D-ResNet50 model, while ``ResNet-S" refers to the 3D-ResNet18 model. Correspondingly, ``VIT-S" represents the smaller variant of the video transformer model, while ``VIT-B" indicates the larger variant. Please keep in mind these specific references for clarity in our discussions.
Based on the table presented in Table~\ref{tab:model_comparison_label1}, the two best-performing architectures on the UCF-101 (1\%) and Kinetics-400 (1\%) datasets under fully supervised settings are ResNet-B and VIT-S. Here's a critical analysis of the motivation provided based on these results:

\textbf{Performance of VIT-S:} Table~\ref{tab:model_comparison_label1} indicates that VIT-S achieves competitive performance compared to ResNet-B, especially on the UCF-101 (1\%) dataset. This suggests that VIT-S is effective in capturing temporal dependencies, as highlighted in the motivation. The transformer-based architecture of VIT-S enables it to capture long-range dependencies efficiently, which is crucial for understanding temporal dynamics in videos.

\textbf{Comparison with ResNet-S:} While ResNet-S also performs reasonably well, especially considering its simplicity compared to VIT-S, it falls slightly short in performance compared to VIT-S on both datasets. This reinforces the notion that VIT-S's strength in handling temporal dependencies contributes to its superior performance in this context.

\textbf{Comparison with VIT-B:} On the other hand, VIT-B exhibits lower performance compared to VIT-S on both datasets. This can be attributed to VIT-B's higher parameter count and complexity. In low data regimes like the 1\% dataset used here, the increased capacity of VIT-B may lead to challenges such as underfitting or difficulty in learning meaningful representations, as mentioned in the motivation. This indicates that while VIT-B has the potential to perform well, it may struggle with limited data availability. Additionally, it's worth noting that smaller models, such as VIT-S, tend to perform better in the temporal domain, which aligns with our motivation.

\textbf{Performance of Resnet-B:} On the other hand, ResNet-B emerges as the best-performing backbone architecture, outperforming both VIT-B and ResNet-S. This superiority can be attributed to ResNet-B's proficiency in capturing spatial features and local dependencies in temporal domain. Therefore, while VIT-S excels in handling long range temporal dependencies and contributes to the overall performance of the ResNet-B and VIT-S combination, ResNet-B's effectiveness in capturing spatial features plays a crucial role in achieving the best performance among all the architectures considered. This underscores the importance of selecting suitable backbone architectures and leveraging their strengths to address different aspects of video understanding tasks.

\subsection{Analysis of Augmentations}
\label{sec:aoa}

\begin{table*}[htbp]
\centering
\begin{tabular}{l|c|c|c|c}
\toprule[0.5mm]
\textbf{Model} & \textbf{Spatial} & \textbf{Temporal} & \textbf{UCF-101(1\%)} & \textbf{Kinetic-400(1\%)} \\ \midrule[0.25mm]
Baseline & - & - & 23.9 & 16.9 \\ 
Spatial-only & \checkmark & - & 25.0 & 17.6 \\ 
Temporal-only & - & \checkmark & 25.2 & 17.8 \\ 
\textbf{Spatial+Temporal} & \textbf{\checkmark} & \textbf{\checkmark} & \textbf{26.1} & \textbf{18.3} \\ \bottomrule[0.5mm]
\end{tabular}
\caption{Analysis of Spatial and Temporal Augmentations. The results are presented for datasets Kinetics-400 and UCF-101, both with a labeling ratio of 1\%.}
\label{tab:augmen}
\end{table*}

The impact of spatial augmentation~\cite{Cubuk_2019_CVPR,bouthillier2016dropout} and temporal augmentation~\cite{yang2020video,singh2021semi} is assessed in Table \ref{tab:augmen}. The baseline denotes the removal of both spatial augmentation techniques and temporal augmentation techniques across all branches. Under this condition, there is a significant decrease in experimental performance. 
However, performance improves when spatial or temporal augmentation is individually applied to the baseline. The best result is achieved when both spatial and temporal augmentation are adapted. Therefore, it is recommended to employ data augmentations in both spatial and temporal domains for optimal results.

\section{Empirical Analysis}\label{section.ea}
\begin{figure*}[t]
\centerline{\includegraphics[width=\linewidth]{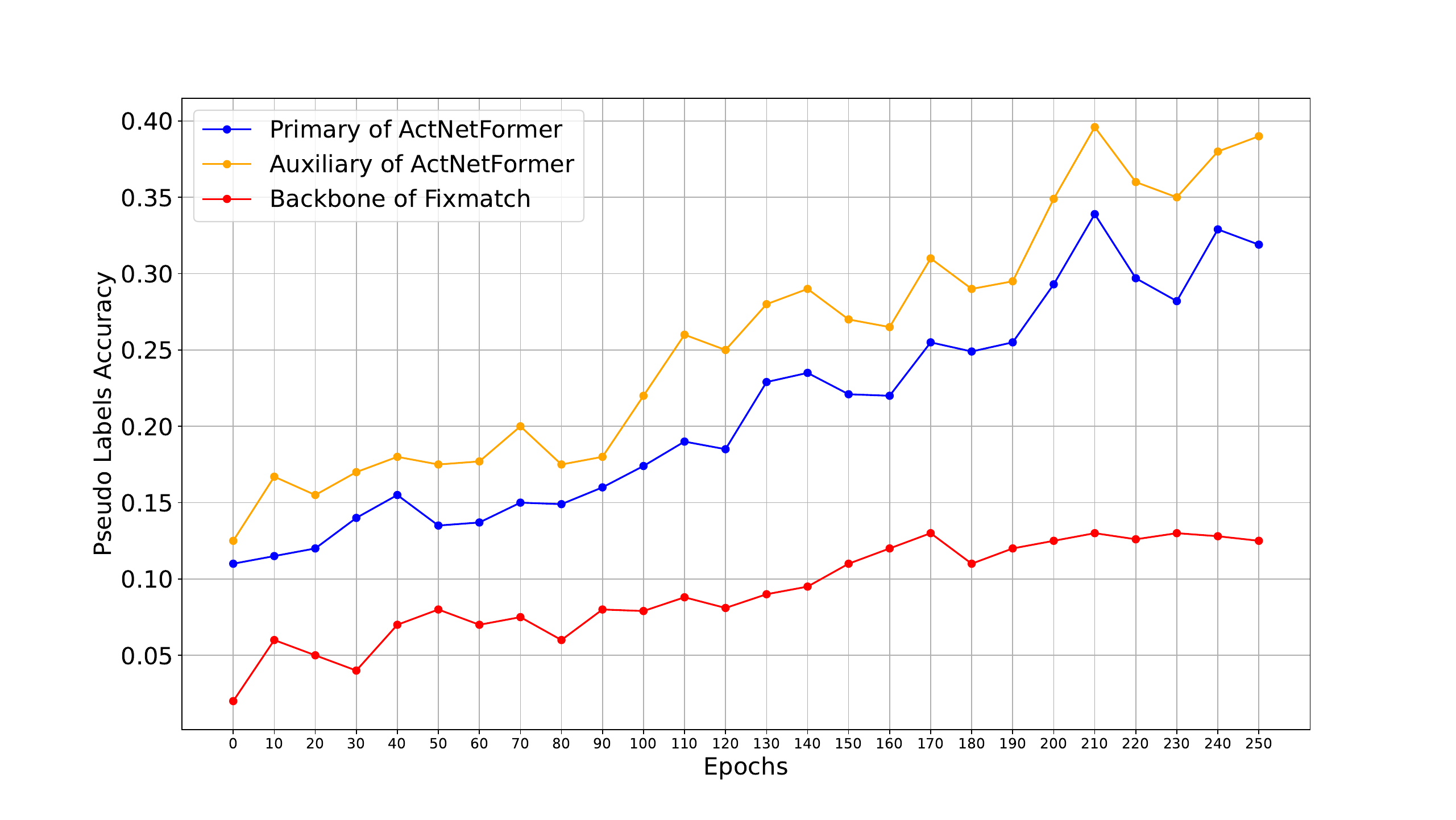}}
\caption{Training accuracy curves illustrate the performance of the primary model of ActNetFormer (\textcolor{blue}{blue}), the model used in FixMatch (\textcolor{red}{red}), and the auxiliary model of ActNetFormer (\textcolor{yellow}{yellow}). Evaluation is conducted on samples assigned pseudo-labels by the auxiliary model.}
 \label{fig:pseudo_label}
\end{figure*}

The impact of auxiliary pseudo-labels is examined in this section. A fundamental distinction between ActNetFormer and FixMatch~\cite{sohn2020fixmatch} lies in the origin of these pseudo-labels. While FixMatch relies solely on its own highly confident predictions, ActNetFormer's primary model (Resnet-B) utilizes pseudo-labels generated by the auxiliary model (VIT-S). To evaluate the quality of pseudo-labels from different sources, we compare three sources: the primary model, the auxiliary model trained in ActNetFormer, and the 3D-ResNet50 trained in FixMatch, which shares the exact similar architecture as the primary model. Initially, we create a subset consisting of samples with pseudo-labels produced by the auxiliary model. Subsequently, the accuracy of these three pseudo-label sources is assessed on this subset, with accuracy recorded at 10-epoch intervals throughout the training process.

Fig.~\ref{fig:pseudo_label} illustrates the accuracy curves. Notably, for samples originally labeled with high confidence by the auxiliary model, the auxiliary model consistently produces superior pseudo-labels (\textcolor{yellow}{yellow curve}) compared to the primary model (\textcolor{blue}{blue curve}) over the training period. Furthermore, in comparison to the architecture learned with FixMatch (\textcolor{red}{red curve}), the primary model gradually improves its estimates for these selected samples. These findings underscore several key points:

\begin{enumerate}
    \item Cross-architecture pseudo-labeling using a different architecture can improve the primary model's performance.
    \item The primary model effectively learns from the auxiliary model through the proposed cross-architecture strategy.
    \item Although in terms of individual performance, the auxiliary model performs less effectively compared to the primary model as shown in Table~\ref{tab:model_comparison_label1}, it still provides useful information to guide the primary model.
\end{enumerate}

\section{Description of the used architecture}

\subsection{Structure of the Primary Model (3D-ResNet50)}

\begin{table}[h]

\centering
\begin{tabular}{m{1cm}m{3.3cm}m{1.5cm}m{2.5cm}}
\toprule[0.5mm]
\textbf{Stage} & \textbf{Block} & \textbf{Stride} & \textbf{Output Size} \\
\midrule[0.25mm]
\addlinespace $input$ & - & - & $224 \times 224 \times 8 \times 3$ \\\addlinespace 
\hline
\addlinespace $conv^1$ & Basic Stem & 1$\times$2$\times$2 & $112 \times 112 \times 4 \times 64$\\\addlinespace 
\hline
\addlinespace $pool$ & Max Pool & 1$\times$2$\times$2 & $56 \times 56 \times 2 \times 64$ \\\addlinespace 
\hline
\addlinespace $res^2$ & $\begin{bmatrix} 1 \times 1 \times 1, 64 \\ 1 \times 3 \times 3, 64 \\ 1 \times 1 \times 1, 256 \end{bmatrix}$ $\times$3 & 1$\times$1$\times$1 & $56 \times 56 \times 2 \times 256 $\\ \addlinespace 
\hline
\addlinespace $res^3$ & $\begin{bmatrix} 1 \times 1 \times 1, 128 \\ 1 \times 3 \times 3, 128 \\ 1 \times 1 \times 1, 512 \end{bmatrix}$ $\times$4 & 1$\times$2$\times$2 & $28 \times 28 \times 1 \times 512$ \\\addlinespace 
\hline
\addlinespace $res^4$ & $\begin{bmatrix} 3 \times 1 \times 1, 256 \\ 1 \times 3 \times 3, 256 \\ 1 \times 1 \times 1, 1024 \end{bmatrix}$ $\times$6 & 1$\times$2$\times$2 & $14 \times 14 \times 1 \times 1024$\\\addlinespace 
\hline
\addlinespace $res^5$ & $\begin{bmatrix} 3 \times 1 \times 1, 512 \\ 1 \times 3 \times 3, 512 \\ 1 \times 1 \times 1, 2048 \end{bmatrix}$ $\times$3 & 1$\times$2$\times$2 & $7 \times 7 \times 1 \times 2048$ \\\addlinespace 
\bottomrule[0.5mm]
\end{tabular}
\caption{Structure of the Primary Model (3D-ResNet50).}
\label{tab:architecturePM}
\end{table}

Table~\ref{tab:architecturePM} shows the architecture of the primary model adapted from~\cite{Feichtenhofer_2019_ICCV}. The network architecture consists of one ResNetBasicStem layer responsible for initial convolution and pooling operations, followed by four ResStage blocks, each containing multiple ResBlock layers for performing the main residual computations. Finally, the architecture concludes with one ResNetBasicHead layer, which performs average pooling, dropout, and linear projection to output the classes. The structure of our primary model which is the 3D-ResNet50 network is characterized by convolutional kernels with dimensions denoted as $\{G_T \times G_H \times G_W, G_C\}$, representing temporal, height, width, and channel sizes respectively. The output size follows the format $\{S^2 \times T \times C\}$, where $C$ denotes the number of channels, $T$ denotes temporal size, and $S$ denotes spatial size. The notation "224 $\times$ 224 $\times$ 3 $\times$ 8 " represents the dimensions of the input video data. "224 $\times$ 224" indicates the spatial resolution of each frame in the video, with both width and height being 224 pixels, "8" represents the number of frames used in a video sequence and "3" signifies the number of channels in each frame, typically representing the Red, Green, and Blue (RGB) color channels in the video.

\subsection{Structure of the Auxiliary Model (VIT-S)}

\begin{figure}[t]
\centerline{\includegraphics[scale=.6]{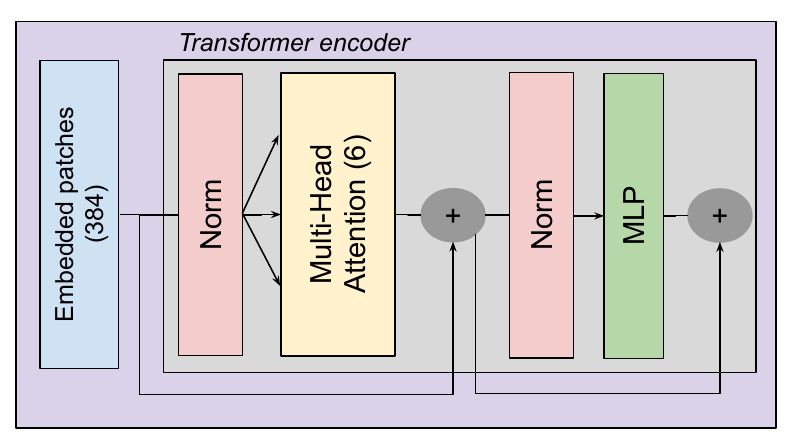}}
\caption{Encoder of the Auxiliary model (VIT-S).}
    \label{fig:architectureAM}
\end{figure}

\begin{table}[t]
  \centering
  \begin{tabular}{m{3.8cm}|m{8cm}}
    \toprule[0.5mm]
    \textbf{Component}         & \textbf{Description}                                                \\ \midrule[0.25mm]
    Model Type                  & Video Transformer (VIT-S)                                                 \\
    Patch Size                  & Input images are divided into patches of size $16 \times 16$       \\
    Number of Layers            & 12                                                                   \\
    Embedding Size              & 384                                                                  \\
    Number of Attention Heads   & 6                                                                   \\
    Dropout Probability         & 0.0                                                                  \\
    Activation Function         & GELU (Gaussian Error Linear Unit)                                  \\
    Normalization               & Layer normalization with $\epsilon=1e-06$                          \\
    Positional Encoding         & Positional encoding is implicit in the patch positions              \\
    Attention Mechanism         & Multi-head self-attention                                           \\
    MLP Hidden Layer Size       & 1536                                                                 \\
    Output                      & Linear layer with output size 101 for (UCF-101) or 400 for (Kinetics-400)   \\ \bottomrule[0.5mm]
  \end{tabular}    
  \caption{Architecture of the auxiliary model (VIT-S).}
  \label{tab:am_summary}
\end{table}

Fig.~\ref{fig:architectureAM} represents the Auxiliary models' video self-attention block, applied on the embedded patches, while Table~\ref{tab:am_summary} summarises the auxiliary model. The auxiliary model (VIT-S) that is used in our ActNetFormer framework is adapted from~\cite{touvron2021training}, designed for processing video data. It incorporates a patch-based approach, where input video frames are divided into patches of size 16x16 pixels using a PatchEmbed layer, facilitating spatial decomposition. Following this, both positional and temporal dropout layers are applied to the embedded patches, aiding in regularizing the model during training.

The core of the model consists of a series of blocks, each of which comprises several components. Within each block, layer normalization is applied to the input embeddings, ensuring stable training dynamics. The self-attention mechanism is employed within each block to capture spatial and temporal relationships from the input video data. This attention mechanism is complemented by multi-layer perceptron (MLP) modules, which enable the model to learn complex non-linear mappings between input and output representations.

Overall, the auxiliary model (VIT-S) leverages the power of self-attention and MLPs to process both spatial and temporal information in video data. The final output of the model is obtained by passing the transformed representations through a linear layer.


\bibliographystyle{splncs04}
\bibliography{bibliography}